# Multi-task Learning for Chinese Word Usage Errors Detection

Jinbin Zhang
School of Data and Computer Science
SUN YAT-SEN UNIVERSITY
Guangzhou, China
e-mail: zhangjb29@mail2.sysu.edu.cn

Heng Wang
School of Data and Computer Science
SUN YAT-SEN UNIVERSITY
Guangzhou, China
e-mail: wangh376@mail2.sysu.edu.cn

*Abstract*—Chinese word usage errors often occur in non-native Chinese learners' writing. It is very helpful for non-native Chinese learners to detect them automatically when learning writing. In this paper, we propose a novel approach, which takes advantages of different auxiliary tasks, such as POS-tagging prediction and word log frequency prediction, to help the task of Chinese word usage error detection. With the help of these auxiliary tasks, we achieve the state-of-the-art results on the performances on the HSK corpus data, without any other extra data.

## I. INTRODUCTION

The Chinese word usage errors (WUEs) often happen in non-native learners' writing. The error may be a semantic error or grammatical error [1]. For example, in a Chinese sentence "我 出出 在 一个 郊区", the Chinese word "出出" is a usage error which needs to be detected, this is a semantic error in Chinese, the true word is "出生". It is an important task to detect the error automatically in the assistant writing system for junior Chinese learners.

There have been some researches focusing on the word error detection and correction with the aid of neural networks (NN). For English, reference [2] applies Bi-LSTM model to deal with the different error types in English. In [3], the authors train the neural network with the task of detecting error in English sentence and the different auxiliary tasks to find the influence of different auxiliary task. Reference [4] adopts a compositional model which is based on neural network to detect grammatical error.

As for Chinese, in [5], the authors adopts Bi-LSTM to detect the grammatical errors. Reference [6] proposes a model to find if there is a WUEs in the Chinese segment. Reference [1] exploits pre-trained word vectors, POS-tag feature and n-gram feature to detect the Chinese word usage errors. There also are many researches focusing on the other type Chinese error, like spelling error [7] [8]and Chinese grammatical error diagnosis [9].

 WUEs detection is a task which is similar to Named Entity Recognition (NER). However, WUEs' labels are very sparse, which is not beneficial to the training of the neural network. One of the purposes involving the multi-task learning in our task is that the auxiliary task with dense labels can alleviate the problem. The similar auxiliary task can help the training of the neural networks. Besides, the multi-task learning can be regarded as a regularizer to reduce overfitting [10].

 Multi-task learning firstly proposed in [11]. Recently, the multi-task learning is becoming popular in multiple machine learning applications. The role of multi-task learning can be seen as implicit data augmentation and regularization [10]. There are many researches combining the multi-task with neural network. Two multi-task learning metods for neural network are soft or hard parameters sharing [10]. There also have been some studies focusing on adopting multi-task learning on different NLP tasks. To name a few, reference [12] involves the language-modeling objective into sequence labeling tasks and achieves significant improvement. Reference [13] discover that "low-level" tasks are better at the lower layers of neural networks. Reference [14] adds the adversarial loss in the multi-task learning for text classification. Reference [15] combines multi-task learning and cross-lingual in their study. Reference [16] utilizes the log frequency objective in the multi-lingual part-of-speech tagging task. Reference [17] explore the contribution of different auxiliary tasks and show that the multi-task is not always effective. Reference [18] incorporate different auxiliary tasks into the keyphrase boundary classification. Reference [19] applies domain adaptation for sequence tagging.

In our paper, we jointly train main task for predicting WUEs' positions and the auxiliary task for predicting word log frequency [16] or predicting POS-tagging, and our model contains a share public part and private part for different tasks. The method of multi-task learning for neural network we adopt is also the hard parameter sharing. We get the state-of-the-art results on the Chinese word usage error detection task. The auxiliary tasks we adopt assists the model in learning better for the main task, compared with no auxiliaries. We also analyze the reason of the different performance on the different auxiliary tasks.

The contributions of our work are as follows,
- We incorporate the multi-task learning into the Chinese WUEs detection task.
- We obtain the state-of-the-art results with the help of auxiliary tasks on the Chinese WUEs detection task.

Our paper is organized as following: our proposed model is shown in Section II, experimental results and analysis in Section III. and We conclude in Section IV.

## II. APPROACH

First, we introduce our model with multi-task learning. It consists of a private part and a public part. The private part is only used for Chinese WUEs detection task. The public part is trained by the WUEs detection task and auxiliary task. Both the private and public part in our model are two Bidirectional Long Short Term Memory (Bi-LSTM) networks [20]. We name the model as **parallel Bi-LSTM**. We first introduce the Bi-LSTM model in the section A.

### A. Bi-LSTM

The LSTM model is a time sequence model, which was proposed in [20]. The LSTM model avoids the long-term dependency problem of the vanilla RNN model [21]. The model has been successfully applied on different applications. For NLP, LSTM model has been adopted for machine translation, sequence labelling, sentence representation and so on. The Bi-LSTM is the concatenation of the backward LSTM and forward LSTM. The Bi-LSTM can exploit the backward and forward information flow compared with the LSTM model, which is perform better on some task. The equations of one direction of the Bi-LSTM model are shown as following,

$$\vec{f}_t = \sigma(W_f \cdot [\vec{h}_{t-1}, x_t]) + b_f \quad (1)$$

$$\vec{i}_t = \sigma(W_i \cdot [\vec{h}_{t-1}, x_t]) + b_i \quad (2)$$

$$\tilde{C}_t = \tanh(W_C \cdot [\vec{h}_{t-1}, x_t] + b_C \quad (3)$$

$$\vec{C}_t = \vec{f}_t \otimes \vec{C}_{t-1} + \vec{i}_t \otimes \tilde{C}_t \quad (4)$$

$$\vec{o}_t = \sigma(W_o \cdot [\vec{h}_{t-1}, x_t]) + b_o \quad (5)$$

$$\vec{h}_t = \vec{o}_t \cdot \tanh(\vec{C}_t) \quad (6)$$

where $W_f, W_i, W_C, W_o, b_f, b_i, b_C$ and $b_o$ are parameters of the Bi-LSTM model, $x_t$ is the input at time t, $\vec{h}_t$ is the hidden representation of forward direction at time t, $\otimes$ is the element-wise product between matrices, $\vec{i}_t$ is the input gate, $\vec{f}_t$ is the forget gate, $\vec{o}_t$ is the output gate, $\tilde{C}_t$ is the cell state at t time. $\sigma$ is the sigmoid function. The different gate control the flow of information at the current time.

The backward direction of the model is similar to the forward direction. The hidden representation of the Bi-LSTM is the concatenation of forward direction and backward direction, we indicate it as,

$$h_t = [\vec{h}_t, \overleftarrow{h}_t] \quad (7)$$

where $h_t$ is the hidden representation of the Bi-LSTM.

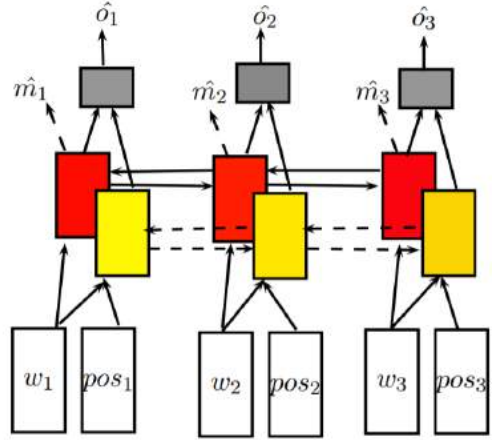

Figure 1. The architecture of our model. The red and yellow boxes represent the public Bi-LSTM network and the private Bi-LSTM network respectively. $w_t$ and $pos_t$ represent the word embeddings and the POS embeddings of the input sentence. $\hat{o}_t$ is the output for main task. $\hat{m}_t$ is the output for auxiliary task. The grey rectangle represents the concatenation of the public Bi-LSTM hidden representation and the private Bi-LSTM hidden representation.

### B. Parallel Bi-LSTM Model

As mentioned above, both the private and public part of our model are Bi-LSTM models. With the former Bi-LSTM shared by both the main task and the auxiliary task, while the latter only used for the main task.

In our model, the input of public Bi-LSTM is word embeddings, while the input of private Bi-LSTM is the concatenation of word embeddings and POS-tagging embeddings. The word embeddings are shared by both two parts. While the POS-tagging embeddings is only used in private Bi-LSTM. For the WUEs position detection task, we use the concatenation of the hidden representation of the private Bi-LSTM network and that of the public Bi-LSTM network.

But for the auxiliary prediction task, only the hidden representation of the public Bi-LSTM is used. Both the private part and the public part are trained together in our experiment.

The procedure shows like this,

Firstly, We concat the hidden representation of public part and private part.

$$h_t^p = [h_t^{pu}, h_t^{pr}] \quad (8)$$

Where $h_t^p$ is the concatenation of the hidden representation of public Bi-LSTM and the hidden representation of private Bi-LSTM at t time. The hidden representation of public Bi-LSTM and private Bi-LSTM are denoted as $h_t^{pu}$ and $h_t^{pr}$ respectively.

The $h_t^p$ are used for predicting WUEs positions,

$$\hat{o}_t = sigmoid(W^e h_t^p + b^e) \quad (9)$$

where $W^e$ and $b^e$ are parameters for predicting WUEs, and $\hat{o}_t$ is the probability of word usage error in the position at time $t$.

TABLE I. THE RESULTS OF MAIN TASK WITH AND WITHOUT AUXILIARY TASKS

| Model | Feature | Accuracy | MRR | Hit@2 | Hit@20% |
|---|---|---|---|---|---|
| Bi-LSTM [1] | CWIN+POS | 0.5138 | 0.6789 | 0.8097 | 0.7479 |
| Bi-LSTM | CWIN+POS | 0.4880 | 0.6600 | 0.6840 | 0.5624 |
| Bi-LSTM | Word2vec+POS | 0.5233 | 0.6831 | 0.7017 | 0.5931 |
| parallel Bi-LSTM | word2vec+POS | 0.5278 | 0.6854 | 0.7011 | 0.5954 |
| parallel Bi-LSTM | word2vec+POS+POS multi-task | 0.5309 | 0.6876 | 0.7051 | 0.5974 |
| parallel Bi-LSTM | word2vec+POS+ log frequency multi-task | **0.5338** | **0.6891** | **0.7060** | **0.6013** |

The public Bi-LSTM hidden representation $h_t^{pu}$ is shared by both the auxiliary task and main task.

The auxiliary task output shows as following,

$$\hat{m}_t = softmax(W^m h_t^{pu} + b^m) \quad (10)$$

where $\hat{m}_t$ is the prediction of the model for word log frequency or POS-tagging task, $W^m$ is the weight, and the $b^m$ is the bias.

Firstly, we compute the loss for main task, which is predicting position of WUEs.

$$L_{error} = \sum_{i=1}^{N} \sum_{t=1}^{L} L_{BCE}(o_t^i, \hat{o}_t^i) \quad (11)$$

Where L is the length of the sentence, and N is the number of sentences.

The loss function we use for main task is binary cross entropy, for one single example, the loss is defined as

$$L_{BCE}(y, \hat{y}) = -\frac{1}{L} \sum_{i=1}^{L} (y_i * \log(\hat{y}_i) + (1 - y_i) * \log(1 - \hat{y}_i)) \quad (12)$$

where L is the length of the sentence.

And we also get the loss for the auxiliary task, the loss function that we use is cross entropy:

$$L_{aux} = \sum_{i=1}^{N} \sum_{t=1}^{L} L_{CE}(m_t^i, \hat{m}_t^i) \quad (13)$$

Finally, we define the total loss of our model as

$$L = L_{error} + \lambda L_{aux} \quad (14)$$

where $\lambda$ is a hyper-parameter to adjust the influence of the auxiliary task. The overall architecture of our model is shown in the Figure 1.

## III. EXPERIMENTS

After We conduct experiments on the Chinese WUE detection task with different auxiliary tasks. The Chinese usage error dataset we adopt is provided in [1]. The corpus is the HSK corpus, which is built by Beijing Language and Culture University. In our experiment, the training set, validation set, and test set are split the same as [1]. The training set contains 8,408 sentences, and both the validation set and the test set contain 1,051 sentences.

As for the setting of our model, 300-dimensional word embeddings are initialized by the pre-trained word vectors which are generated with the gensim word2vec [22] tool, and the corpus we utilize for training word vectors is SogouCS[1]. To get the POS-tag label of our data, we use the tool pyltp2[2]. The size of POS embeddings is 16. The word log frequency label is proposed in [16]. The label is defined as a = int(log(freqtrain(w))). To get the word log frequency labels, we count the number of different word in the HSK corpus, and we transpose the number into the word log frequency. The advantage of the word log frequency is that the labels don't need any other tool or data. The dimension of the hidden state in Bi-LSTM is 100, and the batch size is 32. The optimizer we adopt is Adam [22], with learning rate 0.001. We implement our model with PyTorch[3]. $\lambda$ is a hyper-parameter in our model, we do experiments with different set of the $\lambda$, and we find it better when $\lambda$ is set 0.01. To eliminate the influence caused by the randomness in our model, we adopt 10 different random seeds for model initialization and average the results as our final result. In the experiments, the validation set is used for selecting the best model in the training procedure. All our experiments runs on a single GPU device NVIDIA TITANX.

### A. Evaluation

The measure to evaluate our model is the same as [1]. A predicted error position is correct only if the position is the same as the ground truth error position. The accuracy is the proportion of the number of predicted correct sentences. For a sentence, the word which the model gives the highest score is marked as the predicted error position. And the mean reciprocal rank (MRR) is defined as

$$MRR = \frac{1}{N} \sum_{i=1}^{N} \frac{1}{rank(i)} \quad (15)$$

where rank(i) is the rank of ground truth for the $i_{th}$ sentence in the predicted results. The Hit@k means the prediction is regarded correct if the rank of ground truth in the predicted position is smaller than k or equal to k. The Hit@r% is similar with the Hit@k, but the k is replaced with $\max(1, \lfloor len * r\% \rfloor)$, where len is the length of the sentence.

### B. Results

We reports the results with and without auxiliary tasks. To compare with former state-of-the-art results, we implement the model of previous research [1], with our pre-trained word vectors and our pos-tagging data fed.

---

[1] http://www.sogou.com/labs/resource/cs.php
[2] https://github.com/HIT-SCIR/pyltp
[3] http://pytorch.org/

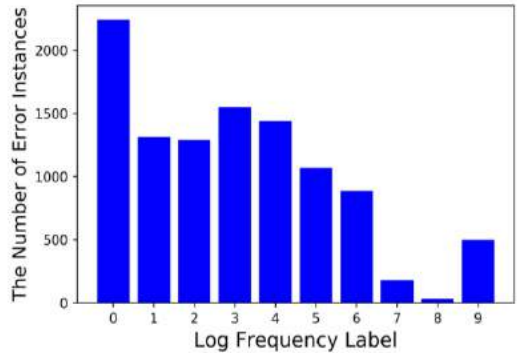

Figure 2. The number of errors on different word log frequency label. The word log frequency is the digit from 0 to 9. The smaller digit represents the lower frequency, while the larger digit represents the higher frequency.

*C. Analysis of the Results*

As the results shown in Table 1, our model with auxiliary tasks achieves state-of-the-art results on Accuracy and MRR. The results of parallel BI-LSTM with the auxiliary task are better than the model without the auxiliary task. The results demonstrate that the auxiliary task we adopt really help the Chinese usage error detection task perform better. We apply two auxiliary tasks in our experiment, and the results indicate that the word log frequency task is better than the POS tagging task. We argue that the POS tagging information has been included in the private part of our model, so the POS tagging task helps less for main task. There are also some researches reporting that the auxiliary task does not always help the main task [14], the auxiliary task should be similar to the main task. To analyze the relation between the word log frequency and WUEs, we count the number of WUEs in the HSK corpus on the different word log frequency label. The statistical result is presented in Figure 2, which shows that the WUEs occur more frequently on the low word log frequency while less on the high word log frequency. This might be the reason why the word log frequency auxiliary task supplements the main task.

IV. CONCLUSION

Our paper introduces a method which incorporates multi-task learning into Chinese word usage error detection task. We perform experiments with different auxiliary tasks like word log frequency prediction task and the POS tagging task, and we obtain the state-of-the-art results with the word log frequency task help. In the future, we will try combining the adversarial training and multi-task learning on the Chinese word usage error detection task.